\documentclass{article}

\usepackage{PRIMEarxiv}
\usepackage{multirow}
\usepackage[utf8]{inputenc} % allow utf-8 input
\usepackage[T1]{fontenc}    % use 8-bit T1 fonts
\usepackage{hyperref}       % hyperlinks
\usepackage{url}            % simple URL typesetting
\usepackage{booktabs}       % professional-quality tables
\usepackage{amsfonts}       % blackboard math symbols
\usepackage{nicefrac}       % compact symbols for 1/2, etc.
\usepackage{microtype}      % microtypography
\usepackage{lipsum}
\usepackage{fancyhdr}       % header
\usepackage{graphicx}       % graphics
\graphicspath{{media/}}     % organize your images and other figures under media/ folder
\usepackage{tabularx}

\title{ADA-YOLO: Dynamic Fusion of YOLOv8 and Adaptive Heads for Precise Image Detection and Diagnosis
}

\author{
  Shun Liu$^{1}$\\
  \texttt{kevinliuleo@gmail.com}\\
  \And 
  Jianan Zhang$^{2}$\\
  \texttt{zjaqifei@ieee.org}\\
  \And 
  Ruocheng Song$^{1}$\\
  \texttt{2021111150@stu.sufe.edu.cn}\\
  \And 
  Teik Toe Teoh$^{3\dagger}$ \\
  \texttt{ttteoh@ntu.edu.sg}\\
  $^{1}$SIME, Shanghai University of Finance and Economics\\
  $^{2}$School of Mathematics, Shanghai University of Finance and Economics\\
  $^{3}$Nanyang Technological University\\
  $^{\dagger}$Corresponding Author
}

\begin{document}
\maketitle

\begin{abstract}{Object detection and localization are crucial tasks for biomedical image analysis, particularly in the field of hematology where the detection and recognition of blood cells are essential for diagnosis and treatment decisions. While attention-based methods have shown significant progress in object detection in various domains, their application in medical object detection has been limited due to the unique challenges posed by medical imaging datasets. To address this issue, we propose ADA-YOLO, a light-weight yet effective method for medical object detection that integrates attention-based mechanisms with the YOLOv8 architecture. Our proposed method leverages the dynamic feature localisation and parallel regression for computer vision tasks through \textit{adaptive head} module. Empirical experiments were conducted on the Blood Cell Count and Detection (BCCD) dataset to evaluate the effectiveness of ADA-YOLO. The results showed that ADA-YOLO outperforms the YOLOv8 model in mAP (mean average precision) on the BCCD dataset by using more than 3 times less space than YOLOv8. This indicates that our proposed method is effective. Moreover, the light-weight nature of our proposed method makes it suitable for deployment in resource-constrained environments such as mobile devices or edge computing systems. which could ultimately lead to improved diagnosis and treatment outcomes in the field of hematology.}\end{abstract}

\section{Introduction}

Object detection techniques have made significant advancements in recent years, enabling the automated identification and localization of anatomical structures, lesions, or abnormalities. Over the years, significant advancements have been made in object detection methods, driven by the availability of large-scale annotated datasets and the development of deep learning techniques. These technologies have shown great potential in improving medical diagnosis and treatment outcomes.

% 介绍两点
% 1. object detection for medicine的大致发展脉络
% 2. 未来值得关注的点
Traditional approaches for object detection in medical imaging relied on handcrafted features and machine learning algorithms prior to the emergence of deep learning. These methods, such as template matching, edge-based detectors, active shape models, and deformable models, were limited by their struggle with complex structures, interclass variability, and limited generalization capabilities \cite{b11}. The advent of convolutional neural networks (CNNs) revolutionized object detection by enabling end-to-end learning of feature representations from raw image data, leading to the development of deep learning-based object detection frameworks with remarkable performance improvements. Region-based approaches like Faster R-CNN \cite{b12} were among the first successful deep learning-based object detection frameworks and have been adapted in the medical imaging domain to detect tumors, organs, and anatomical landmarks \cite{b1,b2,b3,b4,b8,b13}. For example, Wang et al.\cite{b8} developed the ChestX-ray8 database, a large-scale dataset for weakly-supervised classification and localization of thorax diseases using chest X-rays. Their work demonstrated the potential of deep learning-based algorithms in accurately detecting lung nodules with an impressive accuracy of 97.3\%. Gulshan et al.\cite{b2} developed a deep learning algorithm achieving an accuracy of 94\% in detecting diabetic retinopathy. This showcases the potential of object detection technologies as a tool for early detection and diagnosis of this condition. Other potential applications have been explored to some extent. Single-shot approaches like YOLO \cite{b14} and SSD \cite{b15} achieved real-time inference speeds and have been adopted for tasks such as lesion detection, cell detection, and organ localization \cite{b16}. Attention-based methods, which selectively focus on relevant regions of an image to enhance detection accuracy, have recently started gaining attention in medical object detection, leveraging domain-specific knowledge and specifications to improve detection performance \cite{b12}. Transfer learning, involving fine-tuning pretrained models trained on large-scale datasets like ImageNet on medical imaging datasets, has facilitated the development of robust and accurate detectors in various medical applications even with limited training data \cite{b11, b9}.

While object detection technologies hold great promise in medicine, there are challenges that need to be addressed. One such challenge is the requirement of large amounts of annotated data to train deep learning algorithms. Collecting and annotating medical data can be time-consuming and expensive.
Secondly, the interpretability of deep learning algorithms remains a concern, as understanding the decision-making process of these algorithms can be challenging for medical professionals.

% Motivation

% Addressed Problem

This article’s structure is organized in this way. The first section introduces the development of medical images detection technology. The second section delivers the theories and fundamentals behind the proposed ADA-YOLO method. In the third section, several comparative experiments are conducted between the ADA-YOLO method and other methods. The last part summarize the full research and give the prospect of future work.

In brief, the main contributions of this paper are as follows:
\begin{itemize}
\item We propose a light-weight yet powerful object detection model called ADA-YOLO, its effectiveness has been proved by empirical experiments, which can outperform the YOLOv8 in terms of mAP(mean average precision) and also achieve an impressive balance between the precision and recall metrics; Meanwhile, proposed model is memory-efficient in training process, which showcase its great potential in portable scenarios and cloud-edge collaboration.
\item We dive into the intricacies of YOLOv8 architecture, and design an efficient component for object detection called \textit{Adaptive Head(AH)} which incorporates the \textit{Dynamic Visual Feature Localisation (DVF)} and \textit{Joint-Guided Regression Module(JGR)} allowing the parallel computing for bounding box regression and class prediction, enabling more precise object localization and thus enhancing detection accuracy, meanwhile achieve better computational efficiency due to the parallel mechanism.
\item Throughout the extensive experiments, we have demonstrated the presented framework has superior performance in not only multi-class image classification, but also showcase its ability of handling multi-scale object detection tasks, breaking the technical bottlenecks of YOLOx model when faced with small objects with overlaps and class imbalance phenomenon.
\end{itemize}

%%%%%%%%%%%%%%%%%%%%%%%%%%%%%%%%%%%%%%%%%%
\section{Related Work}

\subsection{R-CNN}

Region-based Convolutional Neural Networks, or R-CNNs for short, are a popular class of deep learning models used extensively for object detection in images. The fundamental idea behind the R-CNN architecture is to first identify regions of interest (RoIs) in an image using a selective search algorithm. These RoIs are then inputted into a convolutional neural network (CNN) to extract features. The extracted features are then utilized to classify objects and refine the RoIs. 
The original R-CNN model was proposed by \cite{} and achieved state-of-the-art performance on several object detection benchmarks at that time. However, its computational complexity was a bottleneck because it required running the CNN separately for each RoI, rendering it impractical for real-time applications. To improve efficiency and accuracy, several subsequent works have tackled this limitation. 

Fast R-CNN \cite{b5} enhances the original R-CNN by adding a RoI pooling layer, which facilitates end-to-end training of the network. This means both the classification and bounding box regression are learned jointly, unlike the original R-CNN where these tasks were learned independently. In addition, Fast R-CNN obviates the need to warp RoIs to a fixed size, a significant constraint in the original R-CNN that resulted in poor feature representation.

%%%%%%%%%%%%%%%%%%%%%%
% [Faster RCNN] ##figure
%%%%%%%%%%%%%%%%%%%%%%

% \unskip

Another major improvement to R-CNNs is the Faster R-CNN model presented in 2016 \cite{b12}. Faster R-CNN introduces a Region Proposal Network (RPN) that learns to predict object proposals directly from image features, eliminating the need for a separate selective search algorithm. The RPN is trained to output a set of candidate object bounding boxes, which are used as RoIs for the subsequent CNN. By sharing convolutional features between the RPN and the CNN, Faster R-CNN achieves faster computation and superior performance in terms of accuracy and real-time performance than compared to the previous R-CNN models.   

Apart from Fast R-CNN and Faster R-CNN, various other variants of R-CNN have been created, each with unique contributions. For instance, Mask R-CNN, proposed by Kaiming He et al. in 2017, extends the R-CNN model to also perform instance segmentation. This involves predicting a mask for each object instance in addition to its bounding box. As a result, complex-shaped and occluded object instances can be detected, which is not possible with traditional bounding box detection alone.

%%%%%%%%%%%%%%%%%%%%%%
% % [Mask RCNN] ##figure
% \begin{figure}
% \includegraphics[width=10.5 cm]{img/mask r-cnn}
% \caption{\label{mask r-cnn}}
% \end{figure}   
% \unskip

Cascade R-CNN \cite{b7}, is another variant that uses a cascade of R-CNNs to improve the accuracy of object proposals. The first stage of the cascade R-CNN produces a large number of high-recall but low-precision proposals. These proposals are then refined by subsequent stages to gradually improve both recall and precision.

Despite the success of R-CNNs, several limitations and challenges persist. One of the limitations is their computational requirements. Although Fast R-CNN and Faster R-CNN address this limitation to some extent, they still require significant computation resources, making them unsuitable for real-time applications on low-power devices. To overcome this challenge, researchers are developing more efficient R-CNNs, such as EfficientDet and YOLOv8, which can achieve high accuracy with limited computational resources. Another challenge associated is small object detection. R-CNNs tend to struggle with detecting small objects, which can lead to false negatives or inaccurate bounding boxes. Addressing this challenge requires approaches like image pyramids, which involve scaling the input image at different resolutions, or incorporating multi-scale features to better capture small objects.

\subsection{Yolov8 Model}

\begin{figure}
\centering
\includegraphics[width=14cm]{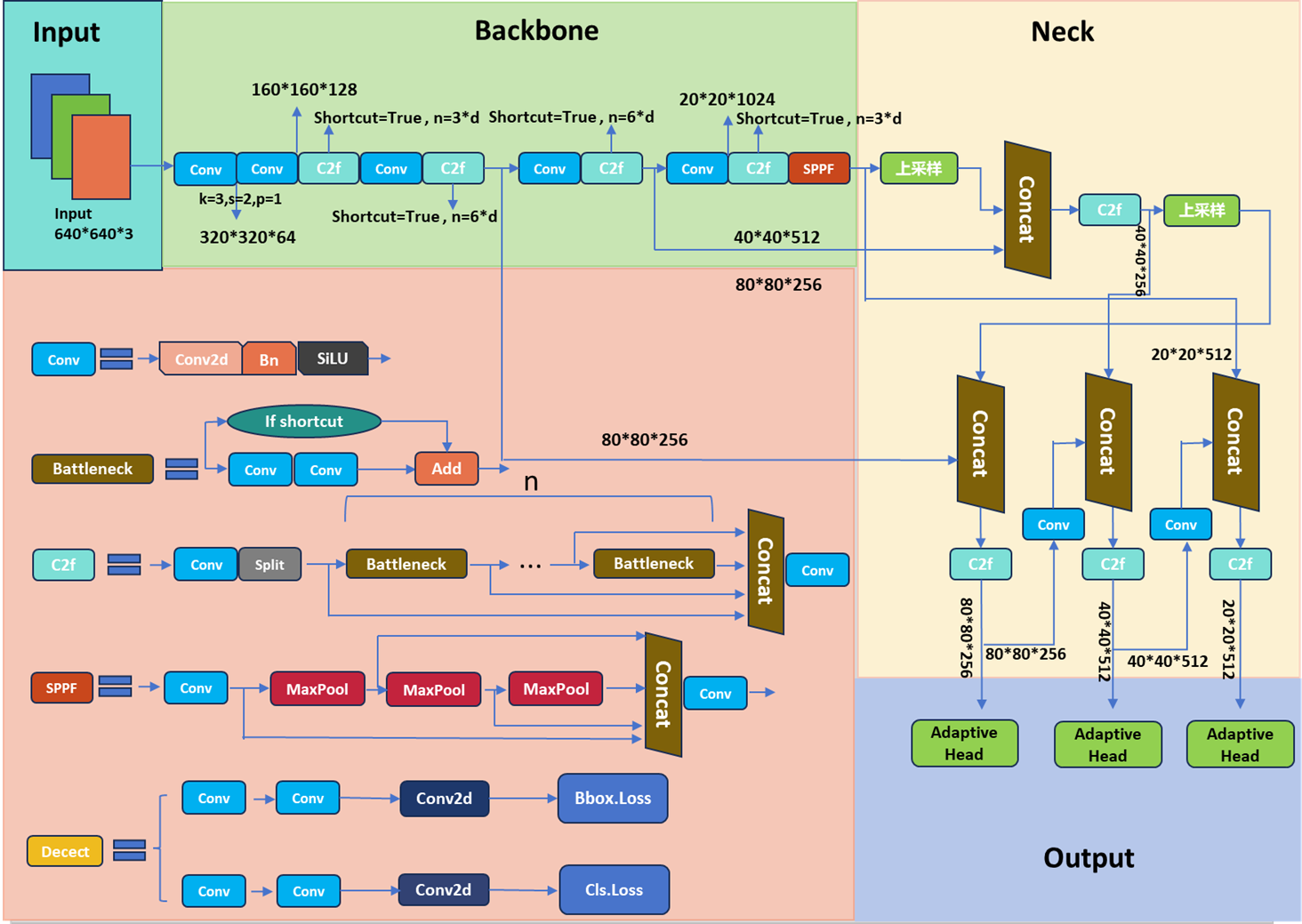}
\caption{Overall architecture of the ADA-YOLO, which is consisted of three components: Backbone for processing, Neck for concatenating and Head(right) for output.\label{yolov8}}
\end{figure}   
% \unskip

\subsubsection{Model Architecture}
YOLOv8:The working principle of YOLOv8 can be summarized as follows: 
The input image is resized into a $448\times 448$ size and fed into the CNN model. 
The CNN model segments the input image into $S\times S$ grids, and each cell is responsible for detecting targets whose center falls within that cell. The core of the CNN consists of multiple convolutional layers, activation functions, and fully connected layers. The convolutional layers are used to perform convolution operations on the input image to extract features. The activation functions introduce nonlinear factors that allow the neural network to better learn and recognize complex features. The fully connected layers are used to connect the convolutional layers with the output layer of YOLO for ultimately identifying the target objects\cite{b19}. 

\textbf{Backbone.}
The backbone in YOLOv8 serves as a feature extractor and is typically constructed using a Convolutional Neural Network (CNN) architecture, such as Darknet, CSPDarknet, or CSPDarknet-tiny, among others. Its primary function is to extract features from the input image, which are utilized for the subsequent tasks like object detection.

Through a series of convolution and pooling operations, the backbone gradually reduces the spatial dimensions of the feature maps while concurrently increasing the depth of these feature maps. This process is designed to capture features of varying hierarchical levels, effectively allowing the neural network to capture and represent features of different complexities and scales within the input image\cite{b17}.

\textbf{Neck.}
The neck is an optional component employed for further processing the feature maps extracted by the backbone. Its primary purpose is to integrate feature information from different hierarchical levels to enhance the performance of object detection. Typically, the neck comprises various operations, such as convolution, upsampling and downsampling, aimed at merging feature maps with different resolutions. This integration is undertaken to improve the network's ability to capture contextual and detailed information about the target objects\cite{b18}.

\textbf{Head.}
The head constitutes a pivotal component in the context of object detection tasks, responsible for the generation of bounding boxes, class probabilities, and object attributes. The head normally consists of a combination of convolutional layers and fully connected layers designed to extract information pertaining to the positions and class labels of the detected objects from the feature maps.The resulting bounding boxes may undergo post-processing steps, such as non-maximum suppression, to obtain the final object detection results\cite{b18}.

\subsubsection{Prediction}
For each cell, YOLO predicts bounding boxes and their confidence scores. These predictions include x, y (predicted center coordinates of the bounding box), w, h (predicted width and height of the bounding box), and confidence (confidence score). Among them, the predicted center coordinates are offset values relative to the upper left corner of the cell, with the unit being the ratio to the size of the cell. The predicted width and height of the bounding box are normalized ratios to the entire image width and height (i.e., w and h are divided by the image w and h, respectively, so that the final w and h fall within the range of 0 to 1).
YOLO treats object detection as a regression problem, allowing a convolutional neural network structure to directly predict bounding boxes and class probabilities from the input image.

\subsubsection{Loss Function} \label{yolo loss}
To better formulate the class imbalance case, the Distribution Focal Loss (DFL)\cite{b41} is ensembled into the original YOLOv8 framework. DFL is an enhanced version of the Focal Loss designed to address the challenge of class imbalance in machine learning tasks, particularly in the context of object detection and classification. The DFL aims to provide a more effective way of handling imbalanced datasets by assigning different weights to individual samples based on their class and adjusting the loss function with a tunable hyperparameter.

The mathematical expression for DFL is given by:

$$
{DFL}(p_t) = -\alpha_t (1-p_t)^\gamma \log(p_t)
$$

In this equation, $p_t$ represents the predicted probability of the target being a positive sample by the model, $\alpha_t$ denotes the weight of class $t$, and $\gamma$ is the tunable hyperparameter. The use of $\alpha_t$ as a weighted factor helps mitigate the effects of class imbalance by assigning higher weights to underrepresented classes.

To address the issue of sample imbalance, $\alpha_t$ is set as an inverse proportion to the number of samples, i.e.,

$$
\alpha_t = \frac{\sum_{i=1}^n w_i}{w_t}
$$

Where $w_i$ represents the weight of the $i$-th sample, $n$ is the total number of samples, and $w_t$ is the weight of class $t$.

Additionally, the parameter $\gamma$ in the DFL formula adjusts the weight of easy and hard samples. When $\gamma > 0$, the weight of easily classifiable samples decreases, while the weight of hard-to-classify samples increases.

In practice, the DFL can be used in combination with other loss functions such as cross-entropy loss to enhance the performance of machine learning models, particularly in scenarios where class imbalances are prevalent, such as object detection in computer vision tasks.
The loss function of the YOLO algorithm mainly includes two parts: classification loss and localization loss. Yolo also introduces a confidence score for predicting the presence of an object and adds a corresponding confidence loss to the loss function
Specifically, the loss function of YOLO can be expressed as:

\begin{equation}
    L_{total} \&= L_{cls} + \lambda_{coord} L_{coord} + \lambda_{noobj} L_{noobj} \\
\end{equation}\label{overall loss}

\begin{equation}
    L_{cls} \&= -\frac{1}{N}\sum_{i=1}^{N}\sum_{j=1}^{S^2}[y_j=i]\log(\hat{y_j}_i) \\
\end{equation}\label{cls loss}

\begin{equation}
    L_{coord} \&= \frac{1}{N}\sum_{i=1}^{N}\sum_{j=1}^{S^2}\sum_{k=1}^{B}[y_j=i](\lambda_{coord}\sum_{l \in \{x,y,w,h\}}[(\hat{t_j})_l - t_j^l]^2) \\
\end{equation}\label{coord loss}

\begin{equation}
    L_{noobj} \&= \frac{1}{N}\sum_{i=1}^{N}\sum_{j=1}^{S^2}\sum_{k=1}^{B}[y_j=0]\hat{p}_{ijk}^2 \\
\end{equation}\label{noobj loss}

 $L_{cls}$ and $\hat y_j$ in Equation \ref{cls loss} denotes category loss and category probability vector of network output respectively. $y_j$ represents the actual tag (YOLO model maps each object to the center of the predicted grid cell, and set label the cell to the object's Category ID) , $N$ and $S$ is the number of samples and the grid size, whilst $B$ indicates the number of bounding boxes predicted for each grid.

$L_{corrd}$ in Equation \ref{coord loss} signifies the location loss, with $\hat t_j$ for bounding box coordinates of network output (including center point coordinates, width, and height) , $t_j$ for bounding box coordinates of real labels, and $lambda_{coord}$ is weight coefficient. 

$L_{noobj}$ displayed in Equation \ref{noobj loss} denotes the confidence loss that is responsible for predicting the existence of an object, $\hat p_{ijk}$ denotes the confidence score of the $k$ bounding box in the $i$-th grid cell of the network output, $[ y_j\ =\ 0 ]$ indicates that there is no indication of an object in the grid cell $j$ and $lambda_{noobj}$ indicates the weight coefficient.

The final overall loss $$L_{total}$$ consists of three parts as shown in Equation \ref{overall loss}: category loss, location loss, and confidence loss, and balances the effects of different loss terms by adjusting $\lambda_{coord}$ and $\lambda_{noobj}$.

\subsection{Ensemble learning}

Ensemble learning improves the generalization performance of a model by combining the predictions of multiple individual models, which can be both homogeneous or heterogeneous and trained with different datasets. The advantage of ensemble learning is that it can leverage the strengths of each model to achieve better performance in terms of prediction accuracy. Additionally, when multiple models are integrated, certain strategies such as voting or weighted sum can be applied to further enhance the performance of the ensemble model.

Before the combination of deep learning and ensemble learning, traditional ensemble learning models mainly used decision trees, support vector machines (SVMs), k-nearest neighbor method (k-NN), etc \cite{b36}. However, these models have certain limitations, such as the inability to express complex nonlinear relationships or handle large-scale datasets effectively, which led to performance limitations in practical applications.

With the development of deep learning, people began to explore combining deep learning models with ensemble learning, hoping to improve the performance of ensemble learning through the good feature learning and fitting ability of deep learning models \cite{b38}. Such deep ensemble learning models can better handle complex nonlinear relationships, express complex patterns, and handle large-scale datasets effectively \cite{b37}.

\begin{itemize}
    \item \textbf{Methods based on model integration.} Bagging and Boosting methods involve training multiple models and combining them. Bagging introduces randomness to reduce overfitting, while boosting adjusts weights to improve robustness; Stacking and methods based on negative correlation: Both methods involve combining multiple base models into a meta-model. Stacking trains multiple base models and uses a meta-model to integrate outputs, while methods based on negative correlation introduce models with negative correlation to reduce variance and improve generalization \cite{b41}.
    
    \item \textbf{Methods based on model complexity.} Explicit integration of neural networks trains multiple neural network models and explicitly combining their outputs. It creates more powerful meta-models, whilst implicit integration of neural networks involves training a single model and creating multiple "sub-models" by fine-tuning its parameters, then combining their outputs. This method can also improve generalization
    
    \item \textbf{Methods based on training data.} Methods based on bagging and boosting refer to the processing of training data. Bagging introduces resampling techniques to randomly sample training data, thereby reducing overfitting. Boosting adjusts weights to focus each model on samples with higher error rates from previous models to generalize better; Stacking and methods based on negative correlation adopt multiple types of models. Stacking introduces multiple base models to form a meta-model, the another create models with negative correlation to reduce variance.
    
    \item \textbf{Methods based on algorithmic aspects.} Boosting and stacking are similar at the algorithmic level. Boosting optimizes algorithm performance through weighted techniques, while Stacking integrates the advantages of different algorithms by combining multiple algorithms into a novel algorithmic structure; Neural network-based methods simultaneously train and adjust neural network models, which includes both explicit and implicit techniques. As mentioned above, explicit techniques involve directly training multiple neural network models and combining them, while implicit techniques involve fine-tuning the parameters of a single neural network model and building on it.

\end{itemize}

% Give a review on Deep Learning Model Ensembling using a TABLE
\begin{table}[ht]
    \caption{Taxonomy of deep ensemble learning, which can be divided into three lines: classical methods, general methods, and fusion strategy.}
    \centering
    \begin{tabular}{|l|l|}
    \hline
        \multicolumn{1}{|c|}{\textbf{Ensemble Strategies}} & \multicolumn{1}{c|}{\textbf{Methods}} \\
    \hline
        \multirow{3}{*}{\textbf{Classical Methods}} & Bagging \\
                                                    & Boosting \\
                                                    & Stacking \\
    \hline
        \multirow{3}{*}{\textbf{General Methods}} & Negative Correlation \\
                                                  & Explicit / Implicit Ensembles \\
                                                  & Homogeneous \& Heterogeneous \\
    \hline
        \multirow{7}{*}{\textbf{Fusion Strategy}} & Unweighted Model Averaging \\
                                                  & Majority Voting \\
                                                  & Bayes Optimal Classifier \\
                                                  & Stacked Generalization \\
                                                  & Super Learner \\
                                                  & Consensus \\
                                                  & Query-By-Committee \\
    \hline
    \end{tabular}
    
    \label{tab:ensemble-strategies}
\end{table}

To make the most of the ensembling, we mostly leverage the stacking method and carry out experiments in Section \ref{lab}.

\section{Methodology}
\subsection{YOLOv8 Object Detection Algorithm}

Figure 3 shows the detailed architecture of YOLOv8. YOLOv8 adopts a backbone network similar to its predecessor YOLOv5\cite{b47}, with notable improvements in the CSPLayer\cite{b27}, now referred to as the C2f module\cite{b19}. The primary objective of this module is to enhance detection accuracy by amalgamating high-level features with contextual information. The YOLOv8 framework employs a single neural network to predict bounding boxes and class probabilities simultaneously, streamlining the object detection process. This characteristic, coupled with its anchor-based mechanism and feature pyramid network, makes YOLOv8 particularly adept at detecting objects efficiently across different scales.

The workflow of YOLOv8 in medical object detection involves several key steps\cite{b28}. Initially, the input medical image is preprocessed to ensure compatibility with the YOLOv8 model. Subsequently, the image passes through the neural network, where feature extraction and object detection occur in a unified manner. YOLOv8 divides the input image into a grid and assigns bounding boxes to each grid cell, predicting object classes and confidence scores. Notably, the use of multiple anchor boxes enhances the model's ability to accurately localize and classify objects of varying sizes\cite{b25}. Post-processing steps, such as non-maximum suppression, further refine the output, ensuring that redundant detections are eliminated, and the most accurate bounding boxes are retained.

In the context of medical object detection, YOLOv8 demonstrates several advantages. Its real-time processing capabilities make it suitable for applications where timely detection is crucial, such as identifying anomalies in medical images. Moreover, the model's ability to handle multi-class detection and its robustness in diverse imaging conditions contribute to its efficacy in medical scenarios with complex visual data. However, it is essential to consider specific challenges, such as the need for specialized datasets and fine-tuning to adapt YOLOv8 to the intricacies of medical imaging. Additionally, ongoing research and collaboration between computer vision and medical experts are pivotal to further refine and optimize YOLOv8 for enhanced performance in medical object detection tasks\cite{b26}.

\subsection{Adaptive Head}
We propose an efficient architecture based on YOLOv8 within an adaptive head finetuned with downstream tasks. In a standard computer vision task, the presented head models the problem as a anchor-based prediction. To dive into the details, detection model utilizes anchor boxes, which are predefined bounding boxes of different sizes and aspect ratios. By predicting offsets and scales for these anchor boxes, the head is able to accurately localize objects of various sizes and shapes. In realistic scenarios, this property can be much beneficial due to the size of input are diversified and stochastic, naive detection method require dozens of image preprocessing or hyper-parameter tuning in order to adapt to the input. Alternatively, our method is subject to an automatic manner to dynamically generates anchor boxes based on the input shape called \textbf{Dynamic Visual Feature Localisation}. This allows the model to adapt to different input resolutions and aspect ratios, making it more versatile and robust in detecting objects in different scenarios \cite{b40}. Different from the \cite{b40}, we drop the activation layer to reserve more information in the intermediate tensors, to better fully capture the transferring semantics, meanwhile increase the computation efficiency to some extent.

Of equal importance, with predefined bounding boxes properties, the detection problem is degraded to a simple regression on scales and offsets, with dynamic anchors settings, the model is able to excavate fine-grained information existed in the images. The architecture also leverage the visual localized features by Dynamic Visual Feature Localisation(DVF) module as shown in the below, to improve the training stability and convergence speed. The DVF module applies adaptive regularization to the predicted bounding box coordinates, allowing the model to learn to scale the predicted boxes appropriately for different object sizes.

Joint-Guided Regression Module(JGRM) is a key component to concurrently calculate the joint score in classification and bounding box tasks. We observe that the performances(or scores) between different target objects are much different, leading to an inductive bias phenomenon, the reason is the separation of class prediction and bounding box regression, this can decrease the robustness of detectors in complex scenario. For example, given an image with various types of cells with different sizes, the model is inclined to detect bounding box in the first place, then fuse the class prediction result, during which the gap generates. In brief, separate training on box regression and prediction is not plausible in the complex detection contexts. Inspired by \cite{b42}, we adopt the distribution focal loss to optimize the regression tasks existed in object detection. It's a refined version of the classic cross entropy function based on the separate training on bounding box and class-wise prediction.

\begin{figure}
    \centering
    \includegraphics[width=14cm]{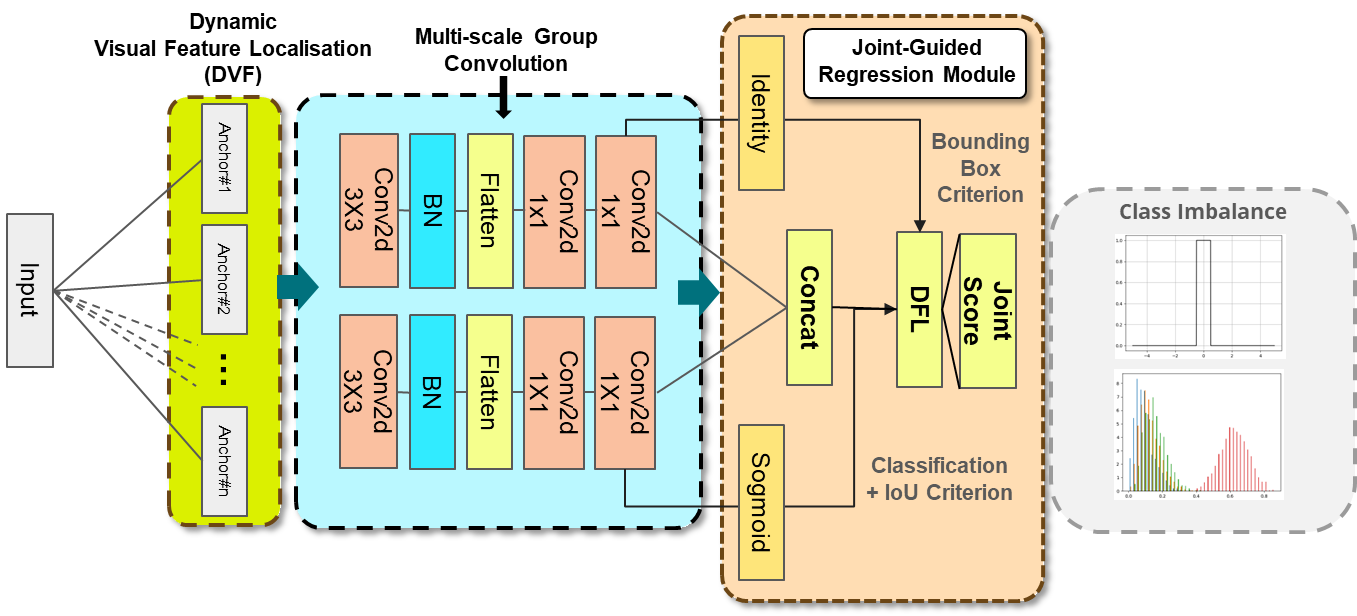}
    \caption{The architecture of the proposed \textbf{\textit{Adaptive Head}}. The refined head uses a streamlined architecture that consists of Dynamic Visual Feature Localisation(DVF), Multi-scale Convolution layers and Joint-Guided Regression Module(JGRM). DVF component focus on the process of automatically resizing the candidate bounding boxes guided by visual features through a regression task; multi-scale convolution layers is proposed to excavate hierarchical and fine-grained information embedded in the images, which is inspired by \cite{b42}; SRM is designed to alleviate the phenomenon of class imbalance where original framework is prone to give more attention to the frequently-visited features, which leave out implicit features, this is very crucial in small object detection and in truncated scenario as well.}
    \label{fig:enter-label}
\end{figure}

The Dynamic Visual Feature Localisation(DVF) as mentioned earlier, aims to give more precise approximation to the image features, the idea is similar to \cite{b44}, which design a sophisticated attention mechanism for detectors, with respect to different feature spatial positions, tasks, etc. In the context of processing a feature tensor \( F \) belonging to a real-valued space with dimensions \( L \), \( S \), and \( C \), we consider the application of self-attention mechanisms. These mechanisms are designed to enhance the feature representation by focusing on relevant aspects of the data.

\textbf{Self-Attention on Feature Tensor.}
The feature tensor \( F \in \mathbb{R}^{L \times S \times C} \) undergoes a self-attention process. This can be represented as the product of the tensor \( F \) and its attention-transformed version, given by:
\begin{equation}
    W(F) = F \times \pi(F),
\end{equation}
where \( \pi(\cdot) \) represents the attention function.

\textbf{Sequential Attention Mechanism.}
To manage the computational complexity, the attention mechanism is divided across the dimensions \( L \), \( S \), and \( C \). This sequential application of attention is expressed as:
\begin{equation}
    W(F) = F \times \pi_C(F \times \pi_S(F \times \pi_L(F))).
\end{equation}
Each function \( \pi_L(\cdot) \), \( \pi_S(\cdot) \), and \( \pi_C(\cdot) \) focuses on its respective dimension.

\textbf{Scale-aware Attention.}
The scale-aware attention mechanism \( \pi_L \) fuses features across different scales. This fusion can be described as:
\begin{equation}
    F \times \pi_L(F) = F \times \sigma\left(f\left({mean}_{S,C}(F)\right)\right),
\end{equation}
where \( f(\cdot) \) is approximated by a \( 1 \times 1 \) convolutional layer and \( \sigma(x) \) is a hard-sigmoid function.

\textbf{Spatial-aware Attention}
The spatial-aware attention \( \pi_S \) aggregates features while focusing on discriminative regions. This aggregation is articulated as:
\begin{equation}
    F \times \pi_S(F) = {mean}_L\left( \sum_{k=1}^{K} w_{l,k} \cdot F(l; p_k + \Delta p_k; c) \times \Delta m_k \right),
\end{equation}
where \( K \) represents the number of sparse sampling locations.

\textbf{Task-aware Attention}
Finally, the task-aware attention \( \pi_C \) dynamically adjusts features for different tasks. This is achieved through channel-wise adjustment:
\begin{equation}
    F \times \pi_C(F) = \max\left(F_c \times \alpha_1(F) + \beta_1(F), F_c \times \alpha_2(F) + \beta_2(F)\right),
\end{equation}
where \( F_c \) is the feature slice at the \( c \)-th channel.The hyper function \( \theta(\cdot) \), crucial for controlling activation thresholds, is denoted by the vector \( [ \alpha_1, \alpha_2, \beta_1, \beta_2 ]^T \). This function, modeled similarly to the approach in reference [3], starts with a global average pooling over the dimensions \( L \) and \( S \) to compress the data dimensions. This is followed by a sequence of two fully connected layers, accompanied by a normalization layer. The process culminates with the application of a shifted sigmoid function that scales the output values to the range \([-1, 1]\).

\section{Experiments} \label{lab}
\subsection{Dataset Description} \label{dataset}
The BCCD dataset \cite{b21} is a comprehensive and informative collection of images containing 12,500 high-resolution microscopic images captured by professional hematologists, the dataset provides a diverse range of blood cell morphology, including normal and abnormal cells, and captures various shapes, sizes, and staining characteristics \cite{b31}. The images represent four major types of blood cells: red blood cells (RBCs), white blood cells (WBCs), platelets, and a combination of these cells. Each image has precise bounding box annotations indicating the location and type of each cell \cite{b30}. 

Meanwhile, corresponding metadata such as cell counts, distributions, and diagnostic information are available when applicable, providing more comprehensive analysis of the data \cite{b28}. The annotations serve as ground truth labels for training and evaluating algorithms, ensuring accurate and reliable labeling. Researchers and practitioners can use this dataset for various applications, including cell counting, cell classification, and anomaly detection. It serves as a benchmark for evaluating the performance of image processing algorithms and machine learning techniques targeted at blood cell analysis. Hence, the BCCD dataset has the potential to advance medical diagnostics, research, and education in the field of hematology \cite{b33}. It enables the development of automated systems that can assist healthcare professionals in quick and accurate blood cell analysis, which can lead improved clinical decision-making and patient care \cite{b29}.

\begin{figure}
  \centering
  \includegraphics[width=0.47\textwidth]{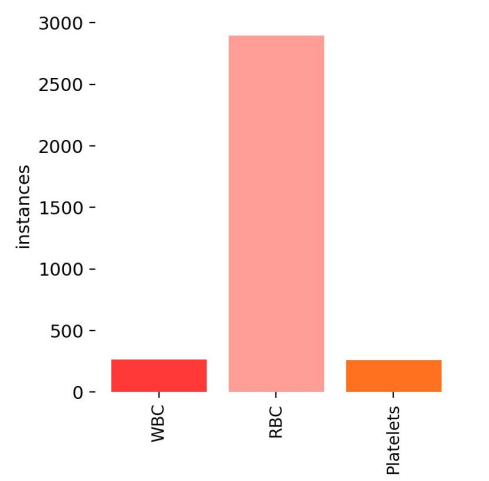}
  \includegraphics[width=0.52\textwidth]{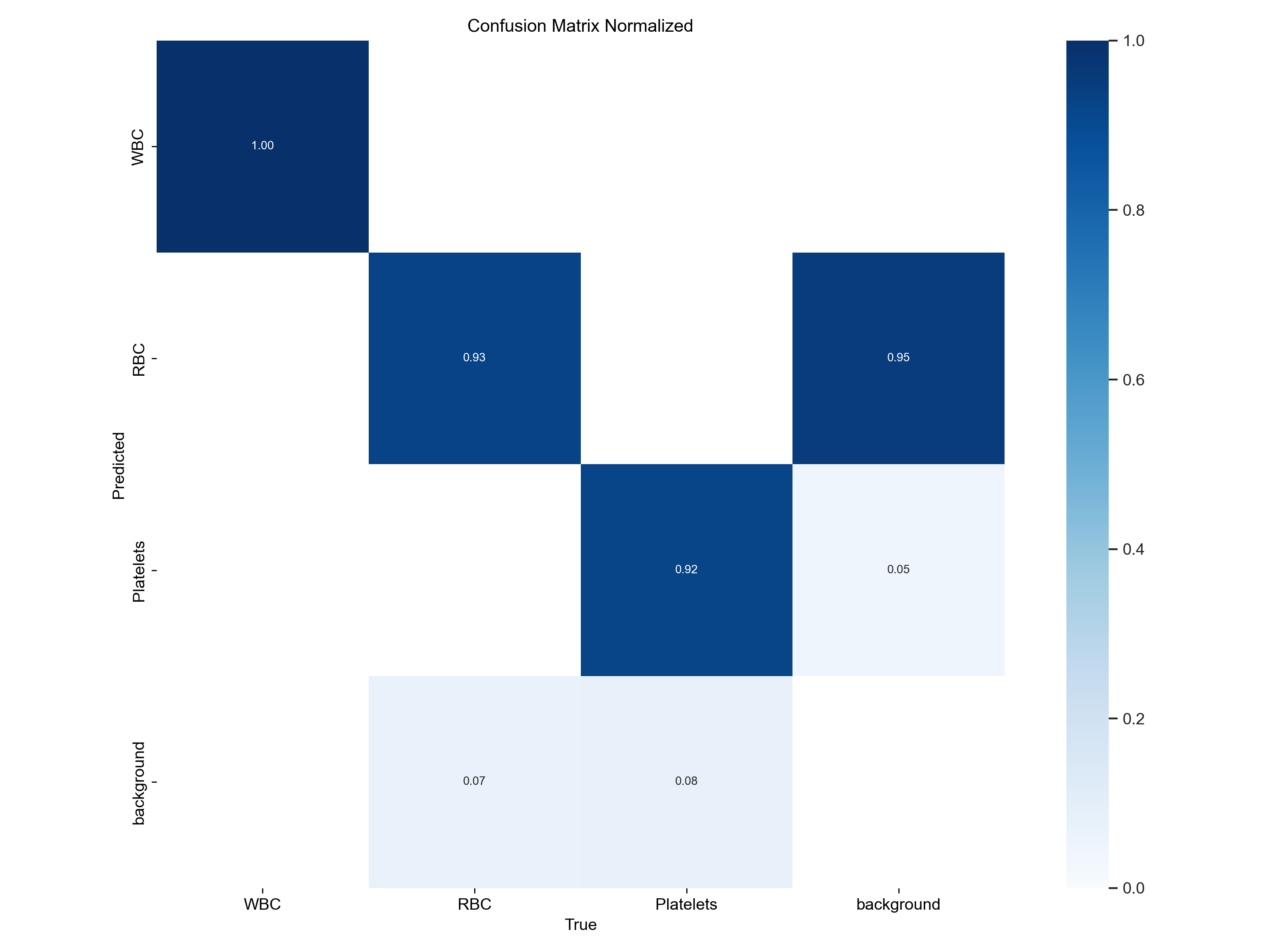}
  \caption{Results of comparative experiments on BCCD dataset}
  \label{distribution}
\end{figure}

Exploring the dataset gives an overview of the model input, this involves visually inspecting a representative sample of images and analyzing the metadata, such as the distribution of cell types and abnormalities.To illustrate the distribution of the classes, we draw a histegram in Figure \ref{distribution}. In the graph, it's apparant that 'RBC' class accounts for the most share with the quantity of approximately 2500, followed 'WBC' class and 'Platelets'

\subsection{Image Preprocessing}
Preprocessing the images is an important step in preparing the dataset for analysis. This includes resizing the images to ensure consistency in image dimensions, normalizing the pixel values to account for variations in lighting and contrast, and considering additional preprocessing techniques such as denoising or histogram equalization based on the experiment's objectives. Preprocessing is necessary to improve the quality of the data and reduce noise to facilitate accurate analysis.

Typically, preprocessing is conducted in the following three ways:

\begin{itemize}
    \item \textbf{Geometric transformations}: This involves operations such as rotation, scaling, cropping, and mirroring to change the size and orientation of the image for training and testing at different scales or angles.
    \item \textbf{Filtering and noise reduction}: Applying various filters and denoising algorithms, such as Gaussian filtering and median filtering, to reduce noise and improve the quality of the image.
    \item \textbf{Color adjustment and contrast enhancement}: This includes adjusting the brightness, contrast, saturation, and hue of the image to enhance its visual appeal and recognizability.
\end{itemize}

\subsection{Experiment Protocols}
The experiments of this paper consists of the hardware and software configurations. Distributed training environment is under Intel 11th-Gen i7-11800H CPU and NVIDIA RTX 3050Ti with the 32.0 GB VRAM(Video Random Access Memory). For software configuration, PyTorch 2.1.0 with cu118 being employed to carry out multi-processing tasks and to make the most of GPUs.

\subsection{Evaluation Metrics}
The evaluation metrics have several components: precision, recall and mAP(mean average precision),.

\textbf{Precision:}
Precision is a measure of the accuracy of positive predictions made by a model. It quantifies the proportion of true positive predictions among all positive predictions made. The formula is given by:

$$
{{Precision}} = \frac{{{{True Positives}}}}{{{{True Positives}} + {{False Positives}}}}
$$
Where True Positives($TP$ for abbreviation) are the correctly predicted positive instances, and False Positives($FP$) are the incorrectly predicted positive instances.

\textbf{Recall:}
Recall, also known as sensitivity or true positive rate, measures the ability of a model to identify all positive instances correctly. It quantifies the proportion of true positive predictions among all actual positive instances. It's represented by a fraction:

$$
{{Recall}} = \frac{{{{True Positives}}}}{{{{True Positives}} + {{False Negatives}}}}
$$
The True Positives and False Positives are exactly the same as the former formula.

\textbf{The Mean Average Precision (MAP):}
MAP is a widely used evaluation metric in information retrieval and machine learning tasks, especially in the field of object detection and image classification \cite{b39}. The MAP indicator measures the quality of ranked lists or retrieval systems, which provides a comprehensive evaluation of the performance of ranking or retrieval systems by considering both precision and recall. Precision measures the accuracy of positive predictions, while recall measures the ability to identify all positive instances. The MAP is obtained by averaging the AP over all classes, providing an overall measure of system quality. To calculate the MAP, we need to compute the Average Precision (AP) for each class or category. AP is then averaged over all classes to obtain the MAP. The formula for AP is given as follows:

$$
{AP} = \frac{1}{{n_{{pos}}}} \sum_{r=1}^{n_{{pos}}} ({precision}(r) \times {recall}(r))
$$

Where n\_{{{pos}}} is the number of positive instances in the ground truth($GT$), and {{precision}}(r) and {{recall}}(r) are the precision and recall values at the r-th retrieved instance. The MAP is calculated as:

$$
{{MAP}} = \frac{{1}}{{n}} \sum_{c=1}^{n} {{AP}}(c)
$$
Where $n$ is the total number of classes or categories.

\subsection{Results}
The experimental results of proposed ADA-YOLO and the baselines(Faster R-CNN \cite{b12}, SSD \cite{b24}, YOLOv5 \cite{b47}, YOLOv7 \cite{b48}, YOLOv8 \cite{b19}).

RT-DETR(Real-Time Detection Transformer) \cite{b23} is a novel target detection method based on Transformer architecture and end-to-end learning, which is different from the traditional target detection method based on region proposal (such as Faster R-cnn) . Rt-detr attempts to make the training and reasoning of target detection simpler and more efficient by completely eliminating manual design components such as anchor frames and non-maximum suppression (NMS). RT-DETR uses self-attention to encode a set of feature vectors globally, and introduces a special"Category embedding" vector to represent locations without targets. During training, RT-DETR minimizes the loss of match between the prediction box and the real box, while trying to match the"No-target" vector to the actual no-target location. This kind of end-to-end learning method makes RT-DETR avoid the super-parameters which need to be adjusted manually in the traditional target detection method, and it is easier to be applied in the target detection task with different sizes and numbers.

\begin{figure}
    \begin{flushright}
      \includegraphics[width=14cm]{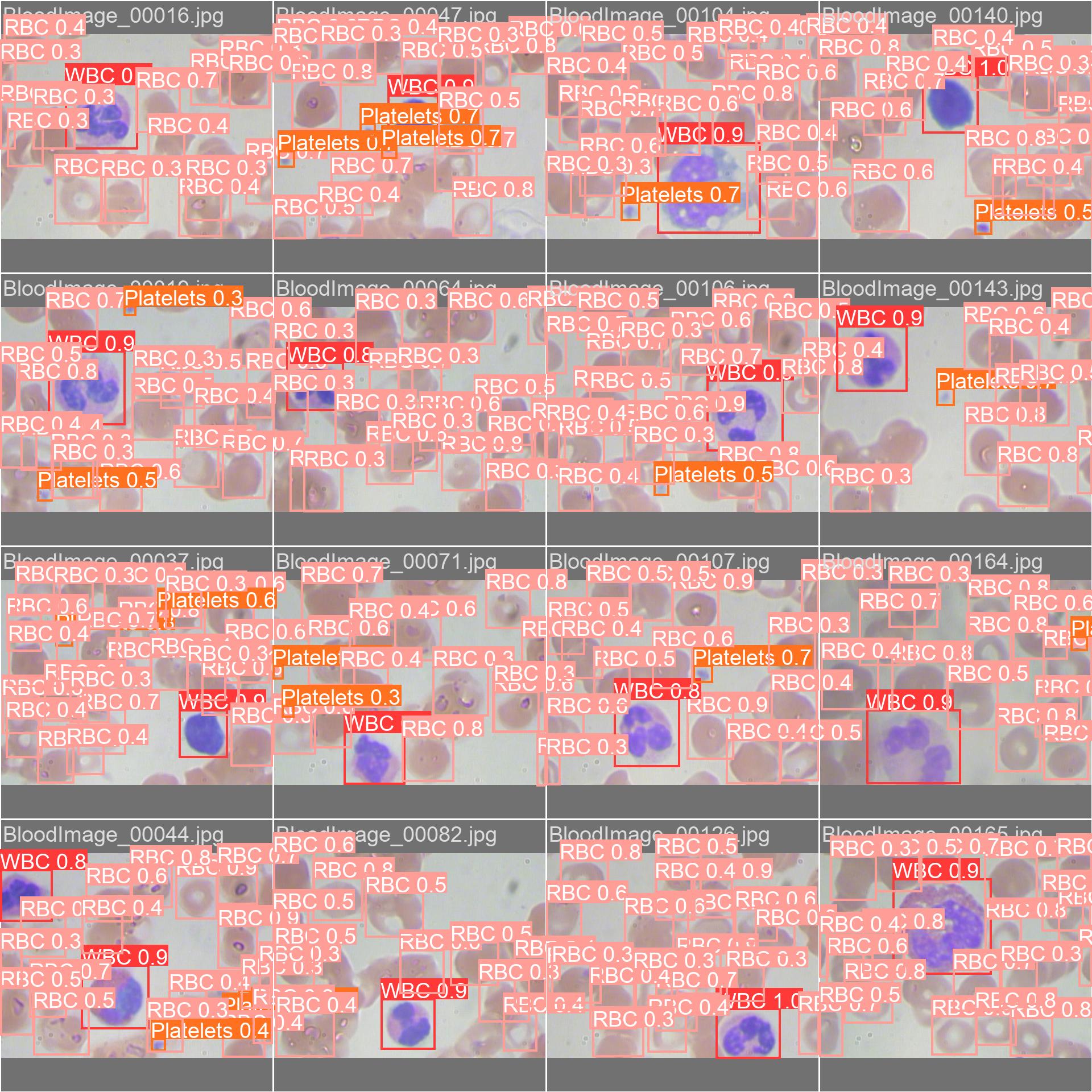}
    \end{flushright}
  \caption{Validation set performance of ADA-YOLO on BCCD dataset, it can be directly infered that ADA-YOLO can successfully detection positive instances of all classes with satisfactory coverage.\label{visualize result}}
\end{figure}

The ADA-YOLO model, proposed for multi-object detection in medical imaging, addresses the challenging issue of object occlusion, or object truncation commonly encountered in medical images. As shown in Figure \ref{visualize result}, proposed methods can successfully detect the missing red blood cells of any sizes, which is a great improvement compared to the baseline. ADA-YOLO has a higher recall rate of 0.918 among all classes, which outperforms the baseline. The result showcases ADA-YOLO's ability to reveal more positive instances in medical object detection. By accurately identifying a higher proportion of true positive cases, even in the presence of occlusions, our model significantly improves disease detection, reduces diagnostic errors, and can translate to earlier and more accurate diagnoses.

% Claims; our method: memory-efficient for scalabel inference framework
Meanwhile, except for remarkable quantitative performance, ADA-YOLO is memory-efficient. To prove this, the main experiment was conducted and it's been shown that YOLOv8 model takes 26.9 MB and calculate 35.1 GFLOPs to predict all samples, whilst our method only need 8.7MB, more than 3 times less space than YOLOv8 and only compute 9.4 GFLOPs. Our model's exceptional efficiency in memory usage is a result of meticulous design and optimization. By strategically implementing compact yet powerful architectural features, we may anticipate the realistic application of this model in portable medical devices, edge computing environments, and resource-constrained healthcare settings, ultimately enhancing accessibility and affordability of advanced medical imaging analysis.

\begin{table}
\centering
\caption{Performances of different methods on BCCD Dataset\label{tab2}}
\begin{tabular}{cccccc}
  \toprule
  \textbf{Methods} & \textbf{mAP@50} & \textbf{mAP@50-95} & \textbf{Recall} & \textbf{Precision} & \textbf{FPS} \\
  \midrule
  RT-DETR \cite{b23} & 0.873 & 0.628 & 0.851 & 0.808 & 35\\
  YOLOv5\cite{b47} & 0.898 & 0.615 & 0.896 & 0.817 & 102\\
  YOLOv7\cite{b48} & 0.907 & 0.615 & 0.890 & 0.845 & 41\\
  YOLOv8 & 0.888 & 0.602 & \textbf{0.903} & 0.819 & 82\\
  \bottomrule
  ADA-YOLO & \textbf{0.912} & \textbf{0.630} & 0.855 & \textbf{0.860} & \textbf{118}\\
  \bottomrule
\end{tabular}
\end{table}

\section{Performances on Other Datasets}
%%%%%%%%%%%%%%%%%%%%%%%%%%%%%%%%%%%%%%%%%%
To further prove the performance of ADA-YOLO, we have conducted grounded experiments on MAR20 dataset \cite{b46}, and visualize the results in Figure \ref{ablation visualize}. Starting with the similar pipeline mentioned in Section \ref{dataset}, we firstly preprocess the input images and then move forward to model training, finally make inference on testing dataset.

The MAR20 object detection dataset \cite{b35} is a comprehensive collection of annotated images designed to facilitate the training and evaluation of object detection algorithms. It encompasses a diverse range of real-world scenarios, including urban streetscapes, natural landscapes, and indoor environments, ensuring the dataset's applicability to various practical settings. Each image in the dataset is annotated with precise location and bounding box information for multiple object classes such as pedestrians, vehicles, and traffic signs. This detailed annotation provides valuable support for algorithm training and evaluation. What unique about it is the inclusion of real-world scenarios enhances the dataset's practical relevance, allowing developers to assess the robustness and generalization capabilities of their algorithms. The MAR20 dataset is suitable for training and evaluating a wide range of object detection algorithms.

\begin{figure}
\centering
    \includegraphics[width=15cm]{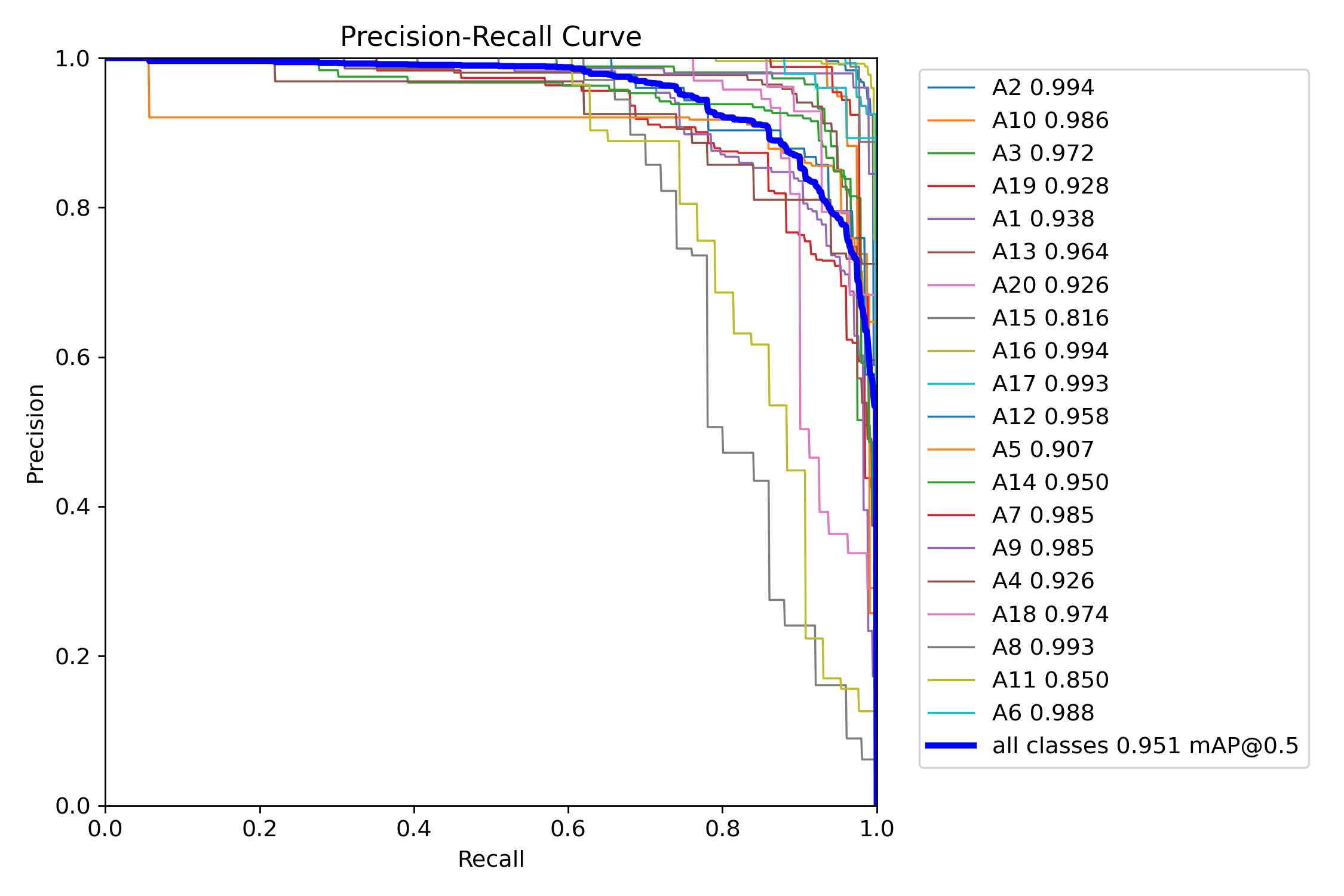}
    \caption{Performance of ADA-YOLO on MAR20 dataset, it can be directly infered that ADA-YOLO can successfully detect positive instances of all classes with satisfactory coverage. When faced with multi-class detection, it can achieve pleasing accuracy.}\label{ablation visualize}
\end{figure}

\begin{figure}
\centering
    \includegraphics[width=10cm]{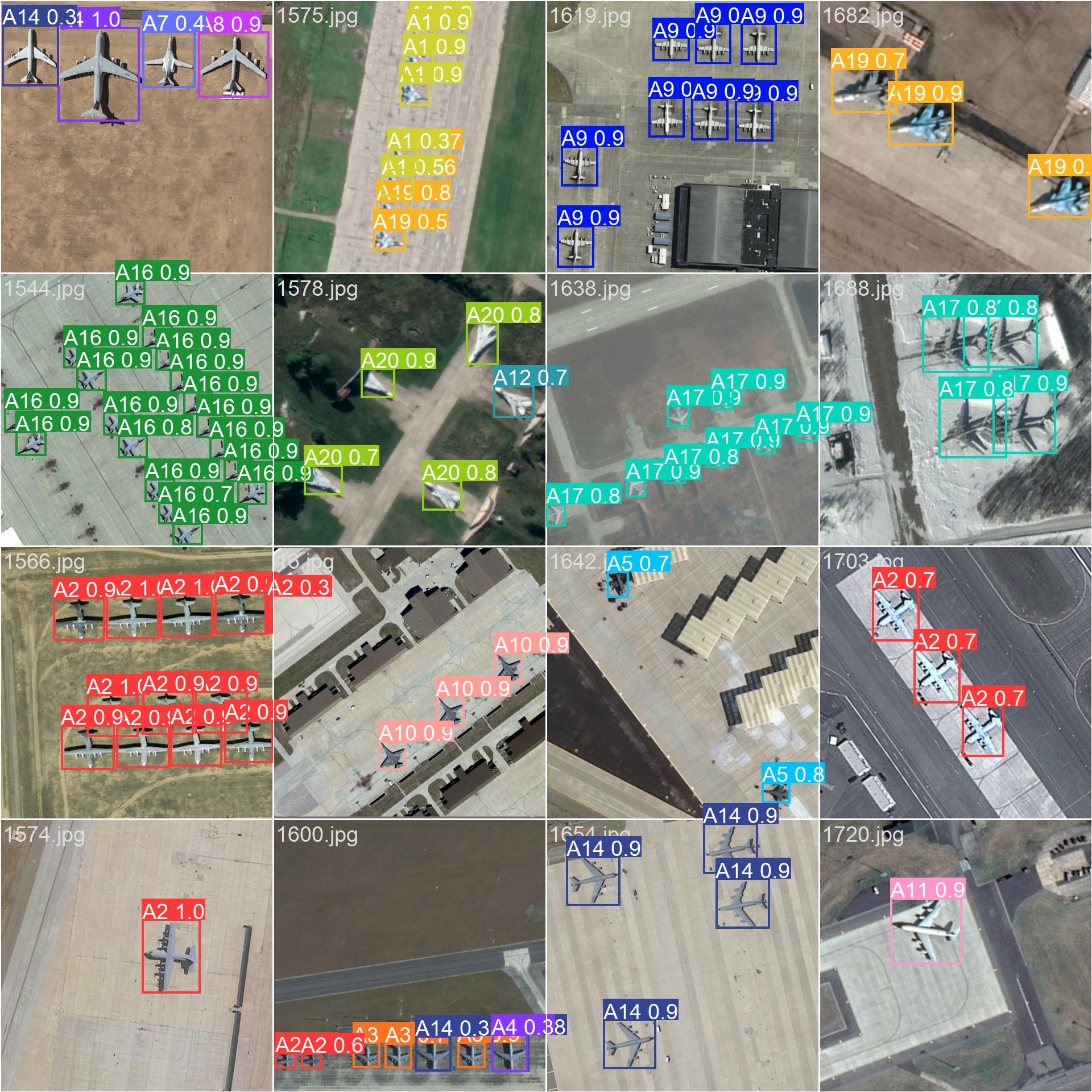}
    \caption{Validation performance of the proposed models on MAR20 dataset.}\label{ablation visualize}
\end{figure}
\begin{figure}
\centering
    \includegraphics[width=10cm]{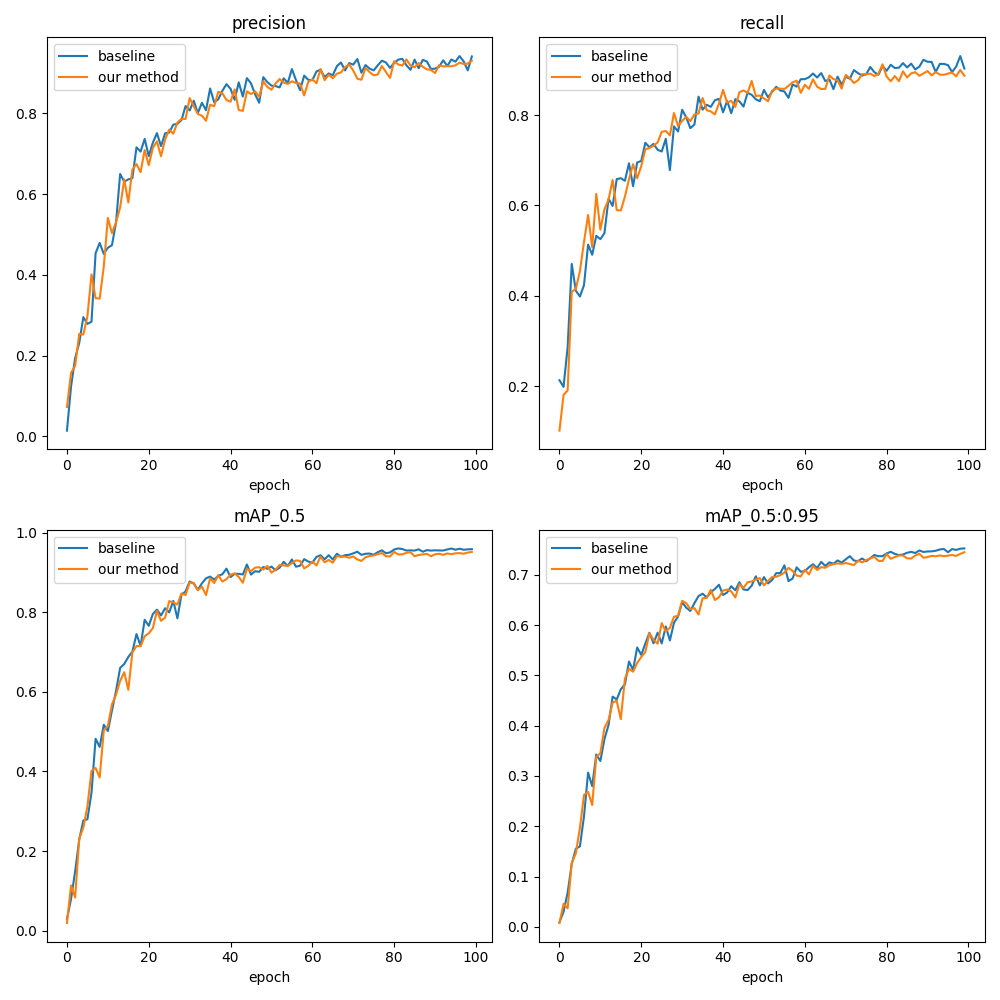}
    \caption{Comparison between ADA-YOLO and YOLOv8 in the training process(left) and validation procedure(right). Our model converges as training epoch increases, the gaps of validation performances between baseline(YOLOv8) and ours narrows down. With much fewer parameters, the results showcases the superiority of the proposed method.}\label{ablation loss}
\end{figure}

In comparison with YOLOv8, our model is adoptable due to its economic momoery consumption and equally-performed properties when dealing with huge dataset. In this experiment, we could observe that ADA-YOLO has the competitive results in MAR20 dataset, with precision of all classes reaching to 85\%, in some classes like A12 the predicted accuracy can even go to 99.4\%. Throughout the results, the strengths of proposed method has been prominent: with fewer model parameters and less memory space to hold, our model can achieve the identical performance regarding of the baselines. It's worth noting that the experiment has the sole group of hyper-parameters, after fine-tuning with the model, we believe its performance can be better, and ultimately surpass the naive method to a larger extent as shown in Figure \ref{ablation loss}(a). As mentioned in Section \ref{yolo loss}, the overall loss is partitioned into three components: bounding box loss, distributed focal loss, and classification loss. From the figure, as the training epochs proceeds to increase, the losses of both model has been consistently decreased and finally converged to some place, whilst our model(the orange line) is within fewer parameters and memory usage. This phenomenon strongly signifies the positive impacts of newly-designed architecture.

From the Figure \ref{ablation loss}(b), we fix the evaluation metrics to be a complex of $P$(precision), $R$(recall), $mAP@50$ and $mAP@50-95$. By this plot, performance gaps between two models is narrowed with the training epochs increasing, which means that proposed model in this experiment setting converges smoothly to the naive method. To the latter half of the trining, ADA-YOLO outperforms the baseline in the all four metrics, showcasing its effectiveness.Through further parameter optimization strategies or feeding with domain-specific knowledge, our model can gain better generality and better performance in corresponding metrics.

\section{Conclusions and Future Work}
In this paper, we propose an memory-efficient yet effective model ADA-YOLO, which leverages a novel architecture and training strategy to enhance the robustness of multi-object detection in medical images. The effectiveness of ADA-YOLO in addressing occlusion-related challenges in medical imaging is demonstrated through the experiments on diverse datasets. The model exhibits superior performance in accurately localizing and classifying multiple objects, even under occlusion conditions, hence it possess a promising future either in computer-aided diagnosis detection due to its innovative design of Adaptive Head. However, this work do have some limitations: Firstly, the huge dataset still cost so much to conduct training and inference, which probably depends on the hardwares in the realistic situation; Secondly, model interpretability can be a potential side effect. By intelligently leveraging in-depth mechanisms and domain knowledge like multi-model embedding, the model can become more reliable and interpretable, paving the way for our future work.

\section{Conclusions and Future Work}
In this paper, we propose an memory-efficient yet effective model ADA-YOLO, which leverages a novel architecture and training strategy to enhance the robustness of multi-object detection in medical images. The effectiveness of ADA-YOLO in addressing occlusion-related challenges in medical imaging is demonstrated through the experiments on diverse datasets. The model exhibits superior performance in accurately localizing and classifying multiple objects, even under occlusion conditions, hence it possess a promising future either in computer-aided diagnosis detection due to its innovative design of Adaptive Head. However, this work do have some limitations: Firstly, the huge dataset still cost so much to conduct training and inference, which probably depends on the hardwares in the realistic situation; Secondly, model interpretability can be a potential side effect. By intelligently leveraging in-depth mechanisms and domain knowledge like multi-model embedding, the model can become more reliable and interpretable, paving the way for our future work.

%Bibliography
\bibliographystyle{unsrt}  
\bibliography{references}

\begin{thebibliography}{10}

\bibitem{b11}
Shaoqing Ren, Kaiming He, Ross~B. Girshick, and Jian Sun.
\newblock Faster r-cnn: Towards real-time object detection with region proposal networks.
\newblock {\em IEEE Trans Pattern Anal Mach Intell}, 39:1137--1149, 2015.

\bibitem{b12}
Fausto Milletari, Nassir Navab, and Seyed-Ahmad Ahmadi.
\newblock V-net: Fully convolutional neural networks for volumetric medical image segmentation.
\newblock {\em International Conference on 3D Vision (3DV)}, abs/1606.04797:565--571, 2016.

\bibitem{b1}
Amy Jin, Serena Yeung, Jeffrey Jopling, Jonathan Krause, Dan Azagury, Arnold Milstein, and Li~Fei-Fei.
\newblock Tool detection and operative skill assessment in surgical videos using region-based convolutional neural networks.
\newblock In {\em 2018 IEEE Winter Conference on Applications of Computer Vision (WACV)}, pages 691--699, 2018.

\bibitem{b2}
Varun Gulshan, Lily Peng, Marc Coram, Martin~C. Stumpe, Derek Wu, Arunachalam Narayanaswamy, Subhashini Venugopalan, Kasumi Widner, Tom Madams, Jorge Cuadros, Ramasamy Kim, Rajiv Raman, Philip~C. Nelson, Jessica~L. Mega, and Dale~R. Webster.
\newblock Development and validation of a deep learning algorithm for detection of diabetic retinopathy in retinal fundus photographs.
\newblock 316:2402--2410, 2016.

\bibitem{b3}
Babak~Ehteshami Bejnordi, Mitko Veta, Paul~Johannes van Diest, Bram van Ginneken, Nico Karssemeijer, Geert Litjens, Jeroen A W~M van~der Laak, Meyke Hermsen, Quirine~F Manson, Maschenka Balkenhol, Oscar Geessink, Nikolaos Stathonikos, Marcory~Crf van Dijk, Peter Bult, Francisco Beca, Andrew~H Beck, Dayong Wang, Aditya Khosla, Rishab Gargeya, Humayun Irshad, Aoxiao Zhong, Qi~Dou, Quanzheng Li, Hao Chen, Huang-Jing Lin, Pheng-Ann Heng, Christian Haß, Elia Bruni, Quincy Wong, Ugur Halici, Mustafa Ümit Öner, Rengul Cetin-Atalay, Matt Berseth, Vitali Khvatkov, Alexei Vylegzhanin, Oren Kraus, Muhammad Shaban, Nasir Rajpoot, Ruqayya Awan, Korsuk Sirinukunwattana, Talha Qaiser, Yee-Wah Tsang, David Tellez, Jonas Annuscheit, Peter Hufnagl, Mira Valkonen, Kimmo Kartasalo, Leena Latonen, Pekka Ruusuvuori, Kaisa Liimatainen, Shadi Albarqouni, Bharti Mungal, Ami George, Stefanie Demirci, Nassir Navab, Seiryo Watanabe, Shigeto Seno, Yoichi Takenaka, Hideo Matsuda, Hady~Ahmady Phoulady, Vassili Kovalev, Alexander
  Kalinovsky, Vitali Liauchuk, Gloria Bueno, M~Milagro Fernandez-Carrobles, Ismael Serrano, Oscar Deniz, Daniel Racoceanu, and Rui Venâncio.
\newblock Diagnostic assessment of deep learning algorithms for detection of lymph node metastases in women with breast cancer.
\newblock {\em Journal of the American Medical Association (JAMA)}, 318.0:2199--2210, 2017.

\bibitem{b4}
Olivier Bernard, Alain Lalande, Clément Zotti, Frederic Cervenansky, Xin Yang, Pheng-Ann Heng, Irem Cetin, Karim Lekadir, Oscar Camara, Miguel Ángel González~Ballester, Gerard Sanroma, Sandy Napel, Steffen~E. Petersen, Georgios Tziritas, Elias Grinias, Mahendra Khened, Alex Varghese, Ganapathy Krishnamurthi, Marc-Michel Rohé, Xavier Pennec, Maxime Sermesant, Fabian Isensee, Paul Jaeger, Klaus~H. Maier-Hein, Peter~M. Full, Ivo Wolf, Sandy Engelhardt, Christian~F. Baumgartner, Lisa~M. Koch, Jelmer~M. Wolterink, Ivana Isgum, Yeonggul Jang, Yoonmi Hong, Jay Patravali, Shubham Jain, Olivier Humbert, and Pierre-Marc Jodoin.
\newblock Deep learning techniques for automatic mri cardiac multi-structures segmentation and diagnosis: Is the problem solved?
\newblock {\em IEEE Transactions on Medical Imaging (TMI)}, 37:2514--2525, 2018.

\bibitem{b8}
Xiaosong Wang, Yifan Peng, Le~Lu, Zhiyong Lu, Mohammadhadi Bagheri, and Ronald~M. Summers.
\newblock Chestx-ray8: Hospital-scale chest x-ray database and benchmarks on weakly-supervised classification and localization of common thorax diseases.
\newblock pages 3462--3471, 2017.

\bibitem{b13}
Joseph Redmon, Santosh~Kumar Divvala, Ross~B. Girshick, and ali farhadi.
\newblock You only look once: Unified, real-time object detection.
\newblock {\em Proceedings of the IEEE conference on computer vision and pattern recognition (CVPR)}, abs/1506.02640:779--788, 2016.

\bibitem{b14}
Wei Liu, Dragomir Anguelov, Dumitru Erhan, Christian Szegedy, and Scott Reed.
\newblock Ssd: Single shot multibox detector.
\newblock {\em European Conference on Computer Vision (ECCV)}, 9905:21--37, 2016.

\bibitem{b15}
H.~R. et~al Roth.
\newblock Ssd: Single shot multibox detector.
\newblock {\em IEEE Transactions on Medical Imaging (TMI)}, 2018.

\bibitem{b16}
Yiting Li, Qingsong Fan, Haisong Huang, Zhenggong Han, and Qiang Gu.
\newblock A modified yolov8 detection network for uav aerial image recognition.
\newblock {\em Drones}, 7, 2023.

\bibitem{b9}
Geert J.~S. Litjens, Thijs Kooi, Babak~Ehteshami Bejnordi, Arnaud Arindra~Adiyoso Setio, Francesco Ciompi, Mohsen Ghafoorian, Jeroen A. W.~M. van~der Laak, Bram van Ginneken, and Clara~I. Sánchez.
\newblock A survey on deep learning in medical image analysis.
\newblock {\em Medical Image Analysis (MIA)}, 42:60--88, 2017.

\bibitem{b5}
Ross~B. Girshick, Jeff Donahue, Trevor Darrell, and Jitendra Malik.
\newblock Rich feature hierarchies for accurate object detection and semantic segmentation.
\newblock {\em Proceedings of the IEEE conference on computer vision and pattern recognition (CVPR)}, 2014:580--587, 2014.

\bibitem{b7}
Zhaowei Cai and Nuno Vasconcelos.
\newblock Cascade r-cnn: Delving into high quality object detection.
\newblock {\em Journal of the American Medical Association}, pages 6154--6162, 2018.

\bibitem{b19}
Xiyang Dai, Yinpeng Chen, Bin Xiao, Dongdong Chen, Mengchen Liu, Lu~Yuan, and Lei Zhang.
\newblock Dynamic head: Unifying object detection heads with attentions.
\newblock {\em Proceedings of the IEEE conference on computer vision and pattern recognition (CVPR)}, abs/2106.08322:7369--7378, 2021.

\bibitem{b17}
Enhui Chai, Lin Ta, Zhanfei Ma, and Min Zhi.
\newblock Erf-yolo: A yolo algorithm compatible with fewer parameters and higher accuracy.
\newblock {\em Image and Vision Computing}, 116:104317--, 2021.

\bibitem{b18}
Juan Terven and Diana Cordova-Esparza.
\newblock A comprehensive review of yolo: From yolov1 to yolov8 and beyond.
\newblock {\em arXiv}, 2023.

\bibitem{b41}
Xue Li, Lifeng Yang, and Xiong Jiao.
\newblock Deep learning-based multiomics integration model for predicting axillary lymph node metastasis in breast cancer.
\newblock {\em Future oncology}, 2023.

\bibitem{b36}
Xibin Dong, Zhiwen Yu, Wenming Cao, Yifan Shi, and Qianli Ma.
\newblock A survey on ensemble learning.
\newblock {\em Front. Comput. Sci}, 14:241--258, 2020.

\bibitem{b38}
Yongquan Yang, Haijun Lv, and Ning Chen.
\newblock A survey on ensemble learning under the era of deep learning.
\newblock {\em Artificial Intelligence Review}, 56:1--45, 2022.

\bibitem{b37}
M.~A. Ganaie, Minghui Hu, A.~K. Malik, M.~Tanveer, and P.~N. Suganthan.
\newblock Ensemble deep learning: A review.
\newblock {\em Engineering Applications of Artificial Intelligence}, 115:105151--, 2022.

\bibitem{b47}
M.~Karthi, V~Muthulakshmi, R~Priscilla, P~Praveen, and K~Vanisri.
\newblock Evolution of yolo-v5 algorithm for object detection: Automated detection of library books and performace validation of dataset.
\newblock {\em International Congress on Shoulder and Elbow Surgery(ICSES)}, pages 1--6, 2021.

\bibitem{b27}
Mohamed Loey, Gunasekaran Manogaran, Mohamed Hamed~N Taha, and Nour Eldeen~M Khalifa.
\newblock Fighting against covid-19: A novel deep learning model based on yolo-v2 with resnet-50 for medical face mask detection.
\newblock {\em Sustainable cities and society}, 65:102600--102600, 2021.

\bibitem{b28}
Xing Wang, Tingfa Xu, Jizhou Zhang, Sining Chen, and Yizhou Zhang.
\newblock So-yolo based wbc detection with fourier ptychographic microscopy.
\newblock {\em IEEE Access}, 6:51566.0--51576.0, 2018.

\bibitem{b25}
Puzhen Wu, Han Weng, Wenting Luo, Yi~Zhan, Lixia Xiong, Hongyan Zhang, and Hai Yan.
\newblock An improved yolov5s based on transformer backbone network for detection and classification of bronchoalveolar lavage cells.
\newblock {\em CSBJ}, 21:2985--3001, 2023.

\bibitem{b26}
Yan Zhou.
\newblock A yolo-nl object detector for real-time detection.
\newblock {\em Expert Systems with Applications}, 238:122256--, 2024.

\bibitem{b40}
Biao Hou, Shenxuan Zhou, Xiaoyu Chen, Heng Jiang, and Hao Wang.
\newblock Yolo-head: An input adaptive neural network preprocessor.
\newblock {\em Intelligence Science}, pages 344--351, 2022.

\bibitem{b42}
Xiang Li, Chengqi Lv, Wenhai Wang, Gang Li, Lingfeng Yang, and Jian Yang.
\newblock Generalized focal loss: Towards efficient representation learning for dense object detection.
\newblock {\em IEEE Transactions on Pattern Analysis and Machine Intelligence}, 45:3139--3153, 2023.

\bibitem{b44}
Xiyang Dai, Yinpeng Chen, Bin Xiao, Dongdong Chen, Mengchen Liu, Lu~Yuan, and Lei Zhang.
\newblock Dynamic head: Unifying object detection heads with attentions.
\newblock {\em arXiv}, abs/2106.08322:7369--7378, 2021.

\bibitem{b21}
Xinyu Liu, Houwen Peng, Ningxin Zheng, Yuqing Yang, Han Hu, and Yixuan Yuan.
\newblock Efficientvit: Memory efficient vision transformer with cascaded group attention.
\newblock {\em CoRR}, abs/2305.07027:14420--14430, 2023.

\bibitem{b31}
Arindam~B. Chowdhury, Jeremy Roberson, Ajat Hukkoo, Srinivas Bodapati, and David~J. Cappelleri.
\newblock Automated complete blood cell count and malaria pathogen detection using convolution neural network.
\newblock {\em IEEE Robot}, 5:1047--1054, 2020.

\bibitem{b30}
Laith Alzubaidi, Mohammed~A. Fadhel, Omran Al-Shamma, Jinglan Zhang, and Ye~Duan.
\newblock Deep learning models for classification of red blood cells in microscopy images to aid in sickle cell anemia diagnosis.
\newblock {\em Electronics}, 9:427--, 2020.

\bibitem{b33}
Nurasyeera Rohaziat, Mohd Razali~Md Tomari, and Wan Nurshazwani~Wan Zakaria.
\newblock White blood cells type detection using yolov5.
\newblock {\em ROMA}, pages 1--6, 2022.

\bibitem{b29}
Zhengfen Jiang, Xin Liu, Zhuangzhi Yan, Wenting Gu, and Jiehui Jiang.
\newblock Improved detection performance in blood cell count by an attention-guided deep learning method.
\newblock {\em OSA Continuum}, 4:323--333, 2021.

\bibitem{b39}
Jeacute;rocirc;me Revaud, Jon Almazaacute;n, Rafael~Sampaio de~Rezende, and Ceacute;sar~Roberto de~Souza.
\newblock Learning with average precision: Training image retrieval with a listwise loss.
\newblock {\em IEEE International Conference on Computer Vision (ICCV)}, abs/1906.07589:5107--5116, 2019.

\bibitem{b24}
Yonglin Wu, Dongxu Gao, Yinfeng Fang, Xue Xu, Hongwei Gao, and Zhaojie Ju.
\newblock Sde-yolo: A novel method for blood cell detection.
\newblock 8, 2023.

\bibitem{b48}
M.~Karthi, V~Muthulakshmi, R~Priscilla, P~Praveen, and K~Vanisri.
\newblock Evolution of yolo-v5 algorithm for object detection: Automated detection of library books and performace validation of dataset.
\newblock {\em Proceedings of the IEEE conference on computer vision and pattern recognition (CVPR)}, pages 1--6, 2021.

\bibitem{b23}
wei liu, dragomir anguelov, dumitru erhan, christian szegedy, and scott reed.
\newblock Ssd: Single shot multibox detector.
\newblock {\em CoRR}, 9905:21--37, 2016.

\bibitem{b46}
Wenqi Yu, Gong Cheng, Meijun Wang, Yanqing Yao, Xingxing Xie, Xiwen Yao, and Junwei Han.
\newblock Mar20: A benchmark for military aircraft recognition in remote sensing images.
\newblock {\em Journal of Remote Sensing (Chinese)}, 2022.

\bibitem{b35}
Shuo Liu, Huanxin Zou, Yazhe Huang, Xu~Cao, Shitian He, Meilin Li, and Yuqing Zhang.
\newblock Erf-rtmdet: An improved small object detection method in remote sensing images.
\newblock {\em Remote Sens}, 15, 2023.

\end{thebibliography}

\end{document}